\documentclass[10pt,twocolumn,letterpaper]{article}

\usepackage{cvpr}
\usepackage{times}
\usepackage{epsfig}
\usepackage{graphicx}
\usepackage{amsmath}
\usepackage{amssymb}

\usepackage{authblk}
\usepackage{adjustbox}
\usepackage{multirow}
\usepackage{booktabs}
\usepackage{float}
\usepackage{bm}
\usepackage{enumitem}
\usepackage[dvipsnames]{xcolor}
\usepackage{flushend}
\usepackage{mathtools}
\usepackage{stmaryrd}
\usepackage{caption}
\usepackage{authblk}
\usepackage{balance}

\usepackage[pagebackref=true,breaklinks=true,colorlinks,bookmarks=false]{hyperref}
   
\cvprfinalcopy % *** Uncomment this line for the final submission

\DeclarePairedDelimiterX\set[1]\lbrace\rbrace{\def\given{\;\delimsize\vert\;}#1}

\def\cvprPaperID{8} % *** Enter the CVPR Paper ID here

\ifcvprfinal\fi

\title{ESL: Entropy-guided Self-supervised Learning \mbox{for Domain Adaptation in Semantic Segmentation}}
\makeatletter
\renewcommand\AB@affilsepx{\qquad\qquad \protect\Affilfont}
\makeatother
\author[1,2]{Antoine Saporta}
\author[1]{Tuan-Hung Vu}
\author[1,2]{Matthieu Cord}
\author[1]{Patrick P\'erez}
\affil[1]{valeo.ai, Paris, France}
\affil[2]{Sorbonne University, Paris, France}

\begin{document}
% \twocolumn[{
% 	\renewcommand\twocolumn[1][]{#1}
% 	\maketitle
% 	\begin{center}
% 		\setlength\tabcolsep{1.5pt}
% 		\footnotesize
% 		\vspace{-0.5cm}
% % 		\begin{tabular}{cccc}
% % 			a) Input image & b) Softmax-based pseudo-labels & c) Entropy-based pseudo-labels & d) Selected in b) but not in c) \\
% % 			\includegraphics[width=.2352\textwidth]{./visualizations/teaser/dusseldorf_000148_000019_leftImg8bit.png} & 
% % 			\includegraphics[width=.2352\textwidth]{./visualizations/teaser/dusseldorf_000148_000019_leftImg8bit_color_ssl.jpg} &
% % 			\includegraphics[width=.2352\textwidth]{./visualizations/teaser/dusseldorf_000148_000019_leftImg8bit_color_esl.jpg} &
% % 			\includegraphics[width=.2352\textwidth]{./visualizations/teaser/dusseldorf_000148_000019_leftImg8bit_color_ssl-esl.jpg} \\
% % 		\end{tabular}
% % 		\captionof{figure}{\small \textbf{Entropy-based pseudo-labels.} The four columns visualize a) RGB input images, b) standard pseudo-labels with maximum softmax predictions, c) our entropy-based pseudo-labels and d) pseudo-labels in b) but excluded by entropy criterion. Most excluded pixels lie on boundary areas where models are most uncertain. \PP{We show that excluding them boost performance of self-trained UDA.}}
% 		\label{lbl:teaser}
% 	\end{center}
% }]
\maketitle
\thispagestyle{empty}
\pagestyle{empty}
%%%%%%%%%%%%%%%%%%%%%%%%%%%%%%%%%%%%%%%%%%%%%%%%%%%%%%%%%%%%%%%%%%%%%%%%%%%%%%%%
\begin{abstract}
While fully-supervised deep learning yields good models for urban scene semantic segmentation, these models struggle to generalize to new environments with different lighting or weather conditions for instance. In addition, producing the extensive pixel-level annotations that the task requires comes at a great cost. Unsupervised domain adaptation (UDA) is one approach that tries to address these issues in order to make such systems more scalable. In particular, self-supervised learning (SSL) has recently become an effective strategy for UDA in semantic segmentation.
At the core of such methods lies `pseudo-labeling', that is, the practice of assigning high-confident class predictions as pseudo-labels, subsequently used as true labels, for target data. To collect pseudo-labels, previous works often rely on the highest softmax score, which we here argue as an unfavorable confidence measurement.

In this work, we propose Entropy-guided Self-supervised Learning (ESL), leveraging entropy as the confidence indicator for producing more accurate pseudo-labels.
On different UDA benchmarks, ESL consistently outperforms strong SSL baselines and achieves state-of-the-art results.
\end{abstract}

%%%%%%%%%%%%%%%%%%%%%%%%%%%%%%%%%%%%%%%%%%%%%%%%%%%%%%%%%%%%%%%%%%%%%%%%%%%%%%%%

\begin{figure*}[t]
    \centering
    %\vspace{4pt}
     \caption{\textbf{Comparison of Softmax-based Self-supervised Learning (SSL) and Entropy-based Self-supervised Learning (ESL)}. While SSL only considers the maximum softmax prediction score, ESL uses the entropy of the output distribution to better assess the confidence of the prediction. Less confident predictions in terms of entropy are excluded from the pseudo-labels by ESL (see right-most figure, best viewed in color and zoomed-in), effectively improving the quality of the label map and boosting the performance of self-trained segmentation networks.}
    \includegraphics[width=\textwidth]{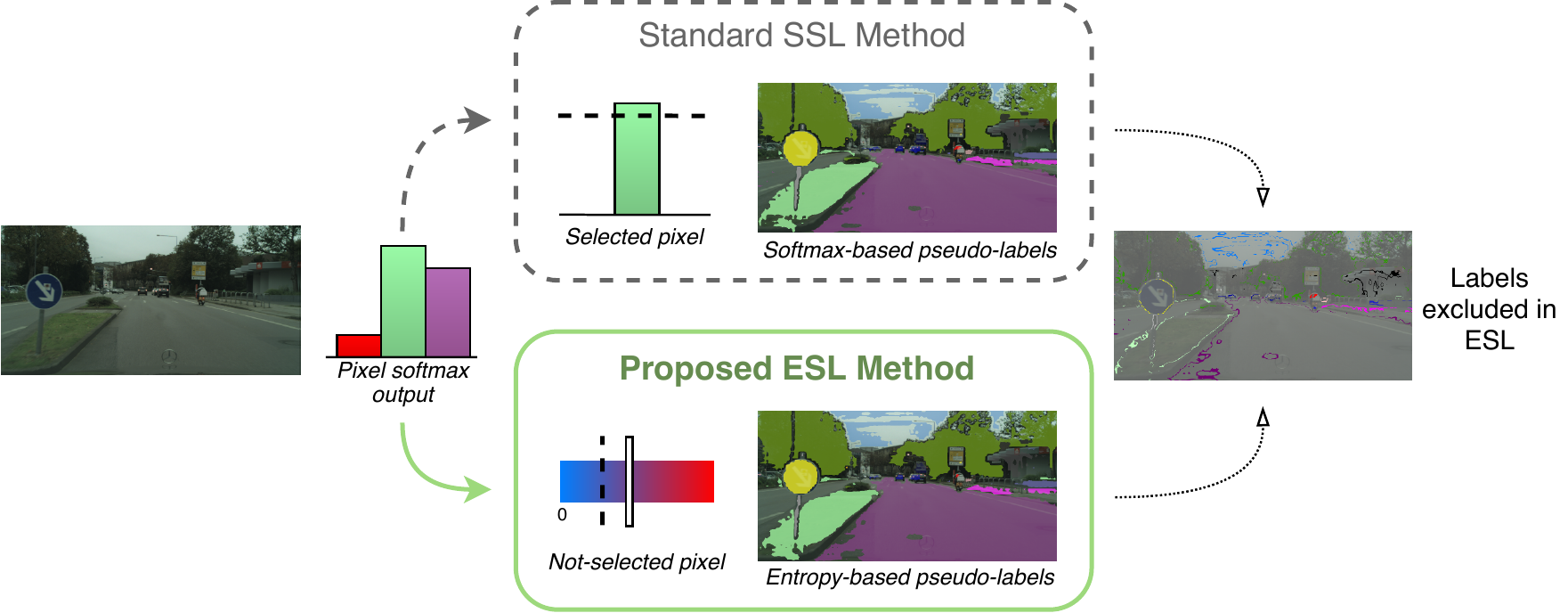}
    %\vspace{-0.8cm}
    \label{fig:teaser}
\end{figure*}

\section{Introduction}
Intelligent systems like autonomous cars often require in-depth understanding of scenes in which they operate.
To this end, most frameworks incorporate semantic segmentation modules to obtain class-label predictions for all input scene pixels.
While recent advances in deep convolutional neural networks (CNNs) have significantly boosted segmentation performance, state-of-the-art results are only achieved with full-supervision. While full supervision already enables perception in functional self-driving vehicles, the annotation cost among others limits the operational domains of such systems without additional techniques to make the learning approach more robust and scalable. To mitigate the indispensable need of manual pixel-level annotations required by the task, recent works are trying to leverage cheaper alternative supervision sources such as synthetic datasets.
However, models solely trained on synthesized images hardly reach comparable performance on real test cases as ones trained on real images.
Such negative effects are originated from the so-called ``domain-gap'' between synthetic and real data, namely \emph{source} and \emph{target}.
To alleviate the performance drop caused by train-test distribution discrepancy, most recent works resort to unsupervised domain adaptation (UDA) techniques.
In the UDA context, along with the annotated source samples, some unlabeled target images are available at train time.
Most UDA works aim at learning domain-invariant representations such that input features to the final classifiers demonstrate insignificant discrepancy across domains.
For that purpose, recent works~\cite{tsai-cvpr2018,vu-cvpr2019} advocate adversarial training as an effective technique.
Orthogonal to adversarial approaches are works motivated by entropy minimization~\cite{vu-cvpr2019}.

Pseudo-labeling is a simple yet effective strategy used in semi-supervised learning~\cite{lee2013pseudo} and recently adopted for domain adaptation~\cite{li-cvpr2019, zou2018unsupervised}.
For UDA, the main idea is to accept high-confident pseudo-labels as if they were true labels of target images at train time.
To that purpose, classes with maximum softmax score are usually selected.
%A common practice is to assign classes having maximum prediction softmax score as pseudo-labels for target samples.
Such a strategy shares the same underlying ``cluster assumption''~\cite{chapelle2005semi} as in entropy minimization methods~\cite{vu-cvpr2019}, i.e., classification decision boundaries should be driven toward low-density regions in the target space.
One major concern of training with pseudo-labels is the lack of guarantee for label correctness, which may eventually cause a ``confirmation bias''~\cite{li2019certainty, tarvainen2017mean}, namely models are biased to previous incorrect pseudo-labels and resist new changes.
To address this problem, one can use a curriculum approach to gradually relax the pseudo-labeling constraint from easy-to-hard~\cite{zou2018unsupervised}.
Self-supervised learning (SSL) done in an iterative manner~\cite{li-cvpr2019} has also been proven effective in improving pseudo-label quality over time.

In this work, we explore the use of entropy, instead of maximum softmax score, as a more reliable criterion for constructing a high-quality set of pseudo-labels.
We argue that maximum softmax score is more error-prone than entropy in cases where the model is highly uncertain.
For example, the right-most image in Fig.~\ref{fig:teaser} visualizes in color the set of pixels that are selected as pseudo-labels based on softmax threshold but do not satisfy the entropy criterion.
Most of these selected pixels lie on the boundaries between object classes, regions where segmentation models are naturally less confident.
By excluding such likely troublesome pixels, we expect to obtain higher-quality pseudo-labels, which indeed mitigates the adverse confirmation bias.
To validate such a hypothesis, starting from the previous state-of-the-art SSL work for UDA, we replace the standard softmax-based pseudo-labeling with our entropy-based strategy.
We coin our framework Entropy-guided Self-supervised Learning (ESL).
%\todo{1-2 lines to describe the self-training iterative strategy + adversarial training based}
On different benchmarks for UDA in semantic segmentation, the proposed ESL scheme shows consistent improvement over strong SSL baselines and achieves state-of-the-art performance.
Extensive experiments and ablation studies bring more insights into the proposed framework.

%\begin{figure}[t!]
%	\centering
%	\includegraphics[width=0.47\textwidth]{teaser.png}
%	\caption{Troublesome pixels where SSL selected but ESL not. \todo{new figure, caption}}
%	\label{lbl:teaser}
%\end{figure}

\section{Related Work}
%\subsection{Unsupervised domain adaptation:}
In last few years, UDA has become an active line of research~\cite{hoffman2017cycada,long2015learning,sun2016deep,tsai-cvpr2018,vu-cvpr2019,wu2018dcan}.
Most works share the similar objective of minimizing distribution discrepancy between source and target domains.
To this end, one can regularize the maximum mean discrepancy (MMD)~\cite{long2015learning}, match correlations of layer activations~\cite{sun2016deep} or adopt adversarial training to align source-target distributions either at feature-level~\cite{hoffman-arxiv2016} or on output space~\cite{tsai-cvpr2018,vu-cvpr2019}.
In some set-ups where extra supervision on the source domain is available at train time, e.g. depth, recent UDA frameworks~\cite{lee-iclr2019, vu-iccv19} leverage such privileged information to further enhance the domain-invariant representations.
For UDA in semantic segmentation, style-transfer-based methods~\cite{huang2017arbitrary,zhu-iccv2017} can be used to translate source images into target-like ones.
The translated source images, with different visual styles yet preserved semantic labels, indeed bring adaptation effects for models trained on them~\cite{hoffman2017cycada,wu2018dcan}.
In this work, we base our iterative self-supervised framework on top of the state-of-the-art adversarial training methods, namely AdaptSegNet~\cite{tsai-cvpr2018} and AdvEnt~\cite{vu-cvpr2019}.

Recent UDA approaches adopt the entropy minimization principle~\cite{vu-cvpr2019} in the attempt to reinforce prediction confidence on the target domain.
Imposing the minimum entropy constraint on target predictions is indeed one way to implement the cluster assumption, namely to prevent decision boundaries from crossing high-density regions.
The pseudo-labeling strategy~in \cite{lee2013pseudo} is in the same spirit.
% Staying under the same spirit is the pseudo-labeling strategy~\cite{lee2013pseudo}.
For UDA in semantic segmentation, Vu \textit{et. al}~\cite{vu-cvpr2019} leverage such a principle by either directly penalizing high-entropy target predictions or, in a more implicit way, aligning source-target distributions on the weighted self-information space with adversarial training.
Different to previous works, we here use entropy as the confidence measurement to construct high-quality sets of pseudo-labels. % feeding to the proposed ESL.

%\subsection{Self-supervised Learning:}
Another efficient UDA approach is self-supervised learning, or self-training~\cite{li-cvpr2019, zou2018unsupervised}.
Such frameworks iterate training upon target pseudo-labels extracted from the preceding step.
For UDA in semantic segmentation, Li \textit{et al.}~\cite{li-cvpr2019}, adopting a simple pseudo-labeling strategy, experimentally demonstrate that the SSL training often converges after two iterations.
With the same methodology, the class-balanced self-training (CBST)~\cite{zou2018unsupervised} solves optimization problems to determine class-specific pseudo-label thresholds while making good use of class-proportion priors.
In our work, we opt to improve over~\cite{li-cvpr2019}, not only because this work reports state-of-the-art performance, but also due its 
superior simplicity. 
%the used pseudo-labeling strategy is much simpler, which indeed curtails unnecessary bell-and-whistle aspects.

\section{ESL: Entropy-guided Self-supervised Learning}
In this Section, we provide details of the proposed ESL framework.
Similar to previous self-supervised works, our learning strategy operates in an iterative manner, in which each iteration involves performing a domain alignment step.
Section~\ref{sec:domain_align_technique} formulates the task and describes alignment techniques that will be used in the experiments.
We then introduce our self-supervised algorithm in Section~\ref{sec:self_training}, followed by details of the entropy-guided pseudo-labeling strategy in Section~\ref{sec:ent_pseudo_labeling}.

\subsection{Domain alignment with adversarial training}\label{sec:domain_align_technique}
%\PPc{Also, make sure there is no verbatim from previous papers.}
In unsupervised domain adaptation, the models are trained with a supervised loss on source domain. Sticking to notations in~\cite{vu-cvpr2019}, we consider a set $\mathcal{X}_s\subset \mathbb{R}^{H\times W \times 3}$ of source training examples coupled with their annotated ground-truth $C$-class segmentation maps $\mathcal{Y}_s \subset [0,1]^{H\times W \times C}$. Each sample $\bm{x}_s$ is a color image of size $H\times W$ and is associated to a map $\bm{y}_s$ that assigns to each pixel $(h,w)$ a one-hot label encoding $\bm{y}_s^{(h,w)} = [\bm{y}_s^{(h,w,c)}]_c$. A semantic segmentation network $F$ takes as input an image $\bm{x}$ and predicts a soft-segmentation map $F(\bm{x})=\bm{P}_{\bm{x}} =[\bm{P}_{\bm{x}}^{(h,c,w)}]_{h,c,w}$ after a final softmax layer. When learning semantic segmentation over the source domain samples $(\bm{x}_s,\bm{y}_s)$, the cross-entropy segmentation loss  % $\mathcal{L}_\text{seg}$ is minimized over the parameters $\theta_F$ of $F$:
\begin{equation}
    \mathcal{L}_\text{seg} (\bm{x}_s, \bm{y}_s) = - \sum\limits_{h=1}^H\sum\limits_{w=1}^W\sum\limits_{c=1}^C\bm{y}_s^{(h,w,c)} \log \bm{P}_{\bm{x}_s}^{(h,w,c)}
    \label{eq:l_seg}
\end{equation}
is minimized over the parameters $\theta_F$ of $F$.

Also available at train time is a set of unlabeled target images $\mathcal{X}_t\subset \mathbb{R}^{H\times W \times 3}$.
At high-level, domain adaptation methods leverage useful statistics from $\mathcal{X}_t$ to learn domain-invariant features, on which the classifier could exhibit similar behaviors regardless of original domains.
Among existing methods, recent ones based on adversarial training~\cite{tsai-cvpr2018,vu-cvpr2019} demonstrate state-of-the-art performance.
In our experiments, we adopt adversarial techniques as the basic block in each of our self-supervised iteration.
It is worth noting though that our overall framework is orthogonal to the distribution alignment technique being used.
%As most state-of-the-art domain adaptation methods use adversarial training for aligning source and target domains, we will use adversarial training notations in what follows. 
%We note that the proposed self-supervised learning techniques could easily be adapted to other domain alignment strategies.\\

In adversarial learning, a discriminator $D$ -- usually, a small fully-convolutional network -- with parameters $\theta_D$ is appended at some layer of $F$ and produces domain classification outputs: output $1$ for the source domain and output $0$ for the target domain. By denoting $\mathcal{L}_\text{adv}$ the cross-entropy loss of this discriminator, the minimization objective of $D$ over $\theta_D$ is:
    \begin{equation}
        \mathcal{L}_\text{D} = \frac{1}{|\mathcal{X}_s|}\sum\limits_{\bm{x}_s\in\mathcal{X}_s}\mathcal{L}_\text{adv}(\bm{x}_s,1) + \frac{1}{|\mathcal{X}_t|}\sum\limits_{\bm{x}_t\in\mathcal{X}_t}\mathcal{L}_\text{adv}(\bm{x}_t,0).
    \end{equation}
    
At the opposite, the semantic segmentation network $F$ must be trained to fool this discriminator $D$. Thus, an adversarial loss is added to the segmentation objective and the training objective of the semantic segmentation network $F$ we minimize over $\theta_F$ is:
    \begin{equation}
        \mathcal{L}_\text{F} = \frac{1}{|\mathcal{X}_s|}\sum\limits_{\bm{x}_s\in\mathcal{X}_s}\mathcal{L}_\text{seg}(\bm{x}_s,\bm{y}_s) + \frac{\lambda_\text{adv}}{|\mathcal{X}_t|}\sum\limits_{\bm{x}_t\in\mathcal{X}_t}\mathcal{L}_\text{adv}(\bm{x}_t,1)
    \end{equation}
with a weight $\lambda_\text{adv}$ for the adversarial term. When training the network, we alternately optimize $D$ and $F$ using the previously defined objective functions $\mathcal{L}_{\text{D}}$ and $\mathcal{L}_{\text{F}}$, respectively.

\subsection{Self-supervised learning}\label{sec:self_training}
On top of the regular domain alignment process, additional self-supervision on the target domain can be added to the objective of the segmentation network $F$ using pseudo-labels on $\mathcal{X}_t$. By noting $(\bm{x}_t,\hat{\bm{y}}_t)$ the pairs of target images and pseudo-labels, the objective function of $F$ with self-supervised learning can be written:
\begin{equation}
    \mathcal{L}_\text{F}^* = \mathcal{L}_\text{F} + \frac{\lambda_\text{SL}}{|\mathcal{X}_t|}\sum\limits_{\bm{x}_t\in\mathcal{X}_t}\mathcal{L}_\text{seg}(\bm{x}_t,\bm{\hat{y}}_t),
\end{equation}
with a weight $\lambda_\text{SL}$ for the self-supervised learning term.

Considering we can already train a segmentation network $F$ by unsupervised domain adaptation without self-supervision, such a semantic segmentation network would give ``good'' predictions in the target domain. Those predictions can serve to collect pseudo-labels for further training. Under this assumption, the self-supervised learning process could be described as follow:
\begin{enumerate}
    \item Train a segmentation network without self-supervision;
    \item Extract pseudo-labels on the target training set using the predictions from the trained network;
    \item Re-train a segmentation network from scratch with additional supervision from the extracted pseudo-labels. %\PPc{from scratch or fine-tuning of the one in 1)?}
\end{enumerate}
This process could be repeated multiple times, extracting finer pseudo-labeling after each iteration. In this work, we will focus on a single iteration of this algorithm. We discuss in what follows ways to extract pseudo-labels on the target training set using a trained network.

\subsection{Entropy-guided Pseudo-label Extraction}\label{sec:ent_pseudo_labeling}

A standard strategy for pseudo-label extraction is to assign classes having maximum prediction softmax score as pseudo-labels. Enforcing this maximum softmax score to be greater than a threshold is a common practice to filter poor predictions from the constructed pseudo-labels. This strategy has already been used in previous state-of-the-art paper in unsupervised domain adaptation for semantic segmentation~\cite{li-cvpr2019}.
Formally, the pseudo-labels extracted using the trained segmentation network $F$ can be written:
\begin{equation}
  \bm{\hat{y}}^{(h, w, c)}_t = 
\begin{cases}
1, & \text{if } \underset{\widetilde{c}}{\arg\max\:} \bm{P}_{\bm{x}_t}^{(h,w,\widetilde{c})} = c\\
  & \text{and } \bm{P}_{\bm{x}_t}^{(h,w,c)} > \mu^{(c)}\\
0, & \text{otherwise}
\end{cases}  
\end{equation}
where $\mu^{(c)}$ is a threshold over the softmax prediction score for class $c$.
Pixels with maximum class score below the relevant threshold are assigned a null pseudo-label vector $\hat{\bm y}_t^{(h,w)} = \boldsymbol{0}$ (not one-hot then). This assignment effectively excludes such pixels from the segmentation loss $\mathcal{L}_\text{seg}(\bm{x}_t,\bm{\hat{y}}_t)$ according to its definition in (\ref{eq:l_seg}).
A simple way to define this threshold, as in~\cite{li-cvpr2019}, would be:
 \begin{equation}
 \resizebox{.485\textwidth}{!}{%
 $\mu^{(c)} = \min\big(\mu^*,  \text{median} \set[\big]{\bm{P}_{\bm{x}_t}^{(h,w,c)} \given \bm{x}_t,h,w \,\wedge\, \underset{\widetilde{c}}{\arg\max\:} \bm{P}_{\bm{x}_t}^{(h,w,\widetilde{c})}=c}\big),$}%
\end{equation}
% \begin{equation}
% \mu^{(c)} = \min\big(\mu^*,  \text{median} \set[\big]{\bm{P}_{\bm{x}_t}^{(h,w,c)}}_{ \bm{x}_t,h,w | \underset{\widetilde{c}}{\arg\max\:} \bm{P}_{\bm{x}_t}^{(h,w,\widetilde{c})} = c} \big) 
% \end{equation}
% \begin{equation}
% \mu^{(c)} = \min(\mu^*,  \text{median} \set[\big]{\bm{P}_{\bm{x}_t}^{(h,w,c)} \given c =  \underset{\widetilde{c}}{\arg\max\:} \bm{P}_{\bm{x}_t}^{(h,w,\widetilde{c})}})
% \end{equation}
where $\mu^*$ is an hyper-parameter. Such a threshold ensures that we keep at least 50\% of the predictions for each class based on their softmax prediction score and that we keep all the predicted labels with a maximum softmax prediction score greater than $\mu^*$ on the easier classes (on which the prediction scores are rather high over the training set).

%\subsection{Entropy-guided Pseudo-label Extraction}
In the standard pseudo-label extraction strategy previously described, the maximum softmax prediction score is used as a confidence score for the prediction. We argue that the maximum softmax prediction score is not the best measure of confidence of the network. Indeed, while the maximum softmax prediction score may be greater than the given threshold, this measure does not take into account the softmax prediction score over other classes, overlooking potentially high softmax prediction score on those other classes that would question the confidence of the model. Such a behavior is illustrated in Figure~\ref{fig:teaser}. As a consequence, we propose an entropy-guided pseudo-label extraction strategy that uses the entropy of the softmax prediction as a measure of confidence. Unlike the maximum softmax prediction score, the entropy actually takes into account the full distribution of the softmax prediction score for each pixel, making this measure more reliable in assessing the confidence of the network. The extracted pseudo-labels can be written as follow:
\begin{equation}
  \bm{\hat{y}}^{(h, w, c)}_t = 
\begin{cases}
1, & \text{if } \underset{\widetilde{c}}{\arg\max\:} \bm{P}_{\bm{x}_t}^{(h,w,\widetilde{c})}=c\\
  & \text{and } \bm{E}_{\bm{x}_t}^{(h,w)} < \nu^{(c)}\\
0, & \text{otherwise}
\end{cases}  
\end{equation}
where the entropy $\bm{E}_{\bm{x}_t}^{(h,w)}$ is defined as:
\begin{equation}
    \bm{E}_{\bm{x}_t}^{(h,w)} = -\: \frac{1}{\log(C)}\sum\limits_{c=1}^C \bm{P}_{\bm{x}_t}^{(h,w,c)} \log \bm{P}_{\bm{x}_t}^{(h,w,c)}
\end{equation}
and $\nu^{(c)}$ is a threshold over the entropy score for pixels of class $c$.
Similarly to the threshold of the previous method, we can define $\nu^{(c)}$ as:
% \begin{equation}
% \nu^{(c)} = \max\big(\nu^*,  \text{median} \set[\big]{\bm{E}_{\bm{x}_t}^{(h,w)}}_{ \bm{x}_t,h,w | \underset{\widetilde{c}}{\arg\max\:} \bm{P}_{\bm{x}_t}^{(h,w,\widetilde{c})} = c} \big), 
% \end{equation}
 \begin{equation}
 \resizebox{.485\textwidth}{!}{%
 $\nu^{(c)} = \max\big(\nu^*,  \text{median} \set[\big]{\bm{E}_{\bm{x}_t}^{(h,w)} \given \bm{x}_t,h,w \,\wedge\, \underset{\widetilde{c}}{\arg\max\:} \bm{P}_{\bm{x}_t}^{(h,w,\widetilde{c})}=c}\big),$}%
\end{equation} 
where $\nu^*$ is a hyper-parameter. 
This threshold ensures we keep at least the 50\% most confident predictions in term of entropy for each class for our pseudo-labels. Moreover, we keep all the predictions with an entropy score lower than the hyper-parameter $\nu^*$ for the easier to predict classes on which the segmentation network is more confident.

\section{Experiments}

In this section, we present experimental results on models trained with self-supervised learning techniques on various semantic segmentation domain adaptation datasets. We compare them to different baselines, showing that self-supervised learning helps boosting the performance and that models trained with entropy-guided self-supervised learning consistently outperform baselines and models with standard self-supervised learning. 

\subsection{Experimental Details}\label{expdet}

\paragraph{Datasets}
In this work, we consider two synthetic source datasets -- SYNTHIA~\cite{ros-cvpr2016} and GTA5~\cite{richter-eecv2016} -- and two real target datasets -- Cityscapes~\cite{cordts-cvpr2016} and Mapillary Vistas~\cite{neuhold-iccv2017}. For SYNTHIA~\cite{ros-cvpr2016}, we use the SYNTHIA-RAND-CITYSCAPES split composed of 9,400 synthetic images of size 1280 $\times$ 760 annotated with pixel-wise semantic labels over 16 classes common with Cityscapes~\cite{cordts-cvpr2016}. GTA5~\cite{richter-eecv2016} is composed of 24,966 synthetic images of size 1914 $\times$ 1052 annotated with 19 classes common with Cityscapes~\cite{cordts-cvpr2016}. For the target datasets, Cityscapes~\cite{cordts-cvpr2016} is a dataset of street-level images split in a training set, a validation set and a testing set. We exclusively use the training set as the target set for domain adaptation. It contains 2,975 images of size 2048 $\times$ 1024. We use the 500-image validation set for testing since ground-truth segmentation maps are missing from the testing dataset. As Cityscapes~\cite{cordts-cvpr2016}, Mapillary Vistas~\cite{neuhold-iccv2017} is a dataset of street-level images split in a training set, a validation set and a testing set. We only use the 18,000 images training set as target set for domain adaptation. The 2,000-image validation set is used for testing because, as for Cityscapes~\cite{cordts-cvpr2016}, the testing set is missing ground-truth segmentation maps.

\paragraph{Architectures and baseline frameworks} We experiment over three state-of-the-art domain adaptation baselines. The three of them are based on DeepLab-V2~\cite{chen-tpami2018} for the semantic segmentation module of the architecture but the domain adaptation frameworks of these baselines are different:
\begin{itemize}
    \item AdaptSegNet~\cite{tsai-cvpr2018} considers semantic segmentations as structured outputs that contain spatial similarities between source and target domains. For this reason, they adopt an adversarial learning framework on the output space. Moreover, they construct a multi-level adversarial network to perform adaptation at different feature levels.
    \item ADVENT~\cite{vu-cvpr2019}, alternatively, adopts an adversarial learning framework on the entropy of the pixel-wise predictions instead of the raw softmax output predictions as in AdaptSegNet~\cite{tsai-cvpr2018}.
    \item BDL~\cite{li-cvpr2019}, as AdaptSegNet~\cite{tsai-cvpr2018}, conducts adaptation on the output space of the semantic segmentation module. The method adds two main components to the framework: first, an image translation module based on CycleGAN~\cite{zhu-iccv2017} to transfer the style of the target domain to source domain images ; second, a bidirectional learning training procedure in which this image translation module and the semantic segmentation module are trained alternately and contribute to each other's performance. Furthermore, this strategy already incorporates self training using standard pseudo-label extraction. In our experiments, we will focus on two given steps of the sequential model training on GTA5 $\rightarrow$ Cityscapes, called `Step 1' and `Step 2', which can be found on the authors' GitHub~\footnote{\url{https://github.com/liyunsheng13/BDL}}. We apply self-training strategies on those two pretrained models, possibly adding image translation (IT).
\end{itemize}

\paragraph{Implementation details}
Implementations are done with the PyTorch deep learning framework~\cite{paszke-nips2017}. Training and validation of the models are done on a single NVIDIA 1080TI GPU with 11GB memory. The semantic segmentation models are initialized with the ResNet-101~\cite{he-cvpr2016} pre-trained on ImageNet~\cite{deng-cvpr2009}. The semantic segmentation models are trained by a Stochastic Gradient Descent optimizer~\cite{bottou2010large} with learning rate $2.5\times 10^{-4}$, momentum $0.9$ and weight decay of $10^{-4}$. The discriminators are trained by an Adam optimizer~\cite{kingma-iclr2015} with learning rate $10^{-4}$. We fix $\lambda_{\text{adv}}$ as $10^{-3}$ and $\lambda_{\text{SL}}$ as 1. Following the recommendation from  \cite{li-cvpr2019}, we use a value of $0.9$ for $\mu^*$ in the SSL experiments. Moreover, we adopt a value of $0.1$ for $\nu^*$ in the ESL experiments and justify this choice in Section~\ref{sec:ablation}. 

\subsection{Results}
\begin{table*}[t]
	\centering
	\vspace{4pt}
	\caption{Comparison of mean intersection-over-union results (in $\%$) for GTA5 $\rightarrow$ Cityscapes experiments.}
	\resizebox{\textwidth}{!}{%
		\begin{tabular}{l|c|c c c c c c c c c c c c c c c c c c c|c}
			\toprule
			\multicolumn{22}{c}{GTA5 $\rightarrow$ Cityscapes}\\
			\midrule
			Method & Self-Training & \rotatebox{90}{road} & \rotatebox{90}{sidewalk} & \rotatebox{90}{building} & \rotatebox{90}{wall} & \rotatebox{90}{fence} & \rotatebox{90}{pole} & \rotatebox{90}{light} & \rotatebox{90}{sign} & \rotatebox{90}{veg} & \rotatebox{90}{terrain} & \rotatebox{90}{sky} & \rotatebox{90}{person} & \rotatebox{90}{rider} & \rotatebox{90}{car} & \rotatebox{90}{truck} & \rotatebox{90}{bus} & \rotatebox{90}{train} & \rotatebox{90}{mbike} & \rotatebox{90}{bike} & mIoU\\
			\midrule
			\multirow{3}{*}{AdaptSegNet~\cite{tsai-cvpr2018}} & - & 79.7 & 16.4 & 76.1 & 18.8 & 12.7 & 24.8 & 33.3 & 20.8 & \textbf{82.0} & \textbf{17.1} & 73.4 & 55.8 & 27.3 & 62.3 & \textbf{37.2} & 30.0 & \textbf{1.4} & 30.8 & 15.1 & 37.6 \\
			& SSL & 80.3 & 17.5 & 78.0 & 19.0 & 19.1 & 26.2 & 36.3 & 22.1 & 81.5 & 17.0 & 72.4 & 55.5 & 28.3 & 62.1 & \textbf{37.2} & 34.9 & 0.9 & \textbf{31.4} & \textbf{20.2} & 38.9\\  
			& ESL & \textbf{81.3} & \textbf{21.7} & \textbf{78.5} & \textbf{20.6} & \textbf{21.2} & \textbf{28.0} & \textbf{37.3} & \textbf{24.8} & 81.1 & 16.1 & \textbf{73.7} & \textbf{56.0} & \textbf{29.1} & \textbf{64.1} & 34.0 & \textbf{35.8} & 0.9 & \textbf{31.4} & 19.2 & \textbf{39.7}\\
			%\midrule
			%\multirow{3}{*}{ADVENT} & Baseline & 88.9 & 18.5 & 80.9 & 25.4 & 26.5 & 28.0 & 33.3 & 18.3 & 83.9 & 34.9 & 78.0 & 57.9 & 29.8 & 84.6 & 30.2 & 44.3 & 0.9 & 26.0 & 24.6 & 42.9\\
			%& SSL & 89.6 & 28.1 & 81.1 & 25.4 & 27.0 & 31.4 & 34.1 & 20.7 & 84.8 & 28.7 & 74.8 & 58.3 & 26.9 & 80.8 & 36.1 & 48.2 & 0.3 & 35.1 & 27.8 & 44.2\\  
			%& ESL & 89.9 & 30.8 & 82.1 & 27.2 & 28.9 & 28.2 & 36.3 & 17.9 & 84.3 & 29.0 & 80.6 & 59.8 & 30.2 & 85.2 & 37.0 & 50.3 & 0.5 & 32.2 & 21.5 & 44.8\\
			\midrule 
			\multirow{3}{*}{ADVENT~\cite{vu-cvpr2019}} & - & 89.9 & 36.5 & 81.6 & 29.2 & 25.2 & 28.5 & 32.3 & 22.4 & 83.9 & \textbf{34.1} & 77.1 & 57.4 & 27.9 & 83.7 & 29.4 & 39.1 & \textbf{1.5} & 28.4 & 23.3 & 43.7\\
			& SSL & 89.6 & 35.4 & 82.0 & \textbf{29.7} & 25.6 & 31.9 & 36.6 & 25.6 & \textbf{84.3} & 29.7 & 75.2 & 59.9 & \textbf{29.9} & 84.7 & \textbf{40.4} & 42.6 & 0.1 & \textbf{32.7} & 30.9 & 45.6\\  
			& ESL & \textbf{90.0} & \textbf{38.6} & \textbf{82.9} & \textbf{29.7}  & \textbf{28.3} & \textbf{33.2} & \textbf{38.5} & \textbf{25.8} & 83.9 & 25.8 & \textbf{78.3} & \textbf{60.0} & \textbf{29.9} & \textbf{85.9} & 35.5 & \textbf{43.3} & 1.1 & 29.1 & \textbf{32.0} & \textbf{45.9}\\ 
			\midrule
			\multirow{5}{*}{BDL~\cite{li-cvpr2019} (step 1)} & - & 88.2 & 41.3 & 83.2 & 28.8 & 21.9 & 31.7 & 35.2 & 28.2 & 83.0 & 26.2 & 83.2 & 57.6 & 27.0 & 77.1 & 27.5 & 34.6 & 2.5 & 28.3 & 36.1 & 44.3\\
			& SSL & 87.3 & 39.4 & 83.7 & 30.2 & 24.9 & \textbf{33.5} & 41.0 & 30.8 & 83.3 & 27.9 & 83.4 & 58.7 & 29.9 & 75.2 & 28.4 & 33.3 & 1.9 & 30.1 & 33.0 & 45.0\\  
			& ESL & 88.4 & 38.6 & 83.9 & 30.4 & 25.9 & 32.8 & \textbf{41.5} & 30.9 & 82.8 & 23.5 & \textbf{85.3} & 59.4 & 30.7 & 78.6 & \textbf{30.8} & 37.3 & \textbf{4.5} & 28.5 & 33.1 & 45.6\\  
			& SSL + IT & 90.2 & \textbf{47.5} & 84.7 & \textbf{33.6} & 26.0 & 33.4 & 39.6 & 33.1 & \textbf{84.1} & \textbf{33.8} & 84.2 & \textbf{59.9} & \textbf{31.5} & \textbf{79.8} & 28.2 & 36.3 & 0.8 & \textbf{31.5} & 31.7 & 46.8\\  
			& ESL + IT & \textbf{90.3} & 46.3 & \textbf{84.8} & 32.7 & \textbf{26.9} & \textbf{33.5} & 39.9 & \textbf{34.8} & 83.9 & 31.2 & 85.0 & 59.2 & 30.3 & \textbf{79.8} & 28.4 & \textbf{43.4} & 1.7 & 28.1 & \textbf{36.2} & \textbf{47.2}\\
			\midrule
			\multirow{5}{*}{BDL~\cite{li-cvpr2019} (step 2)} & - & \textbf{91.0} & \textbf{44.8} & 83.9 & 32.0 & 24.6 & 29.5 & 34.4 & 30.8 & 84.3 & 39.3 & 83.9 & 56.8 & 29.7 & \textbf{83.3} & 35.4 & \textbf{49.8} & 0.2 & 27.3 & \textbf{37.1} & 47.3\\
			& SSL & 89.7 & 39.6 & 84.1 & 30.2 & 28.4 & \textbf{31.9} & 39.0 & 29.4 & 83.9 & 35.1 & \textbf{85.7} & 58.0 & 31.6 & 80.8 & 36.2 & 46.6 & 0.5 & 28.9 & 33.7 & 47.0\\  
			& ESL & 90.0 & 39.2 & 84.3 & 32.0 & \textbf{31.1} & 31.7 & \textbf{39.2} & 32.1 & 83.6 & 31.5 & 84.9 & 58.5 & 31.7 & 82.9 & \textbf{39.5} & 48.4 & 0.9 & 30.5 & 33.0 & 47.6\\  
			& SSL + IT & 90.3 & 43.6 & 84.4 & 32.3 & 28.8 & 31.5 & 37.1 & \textbf{34.2} & \textbf{84.7} & \textbf{42.3} & 84.0 & 58.2 & \textbf{32.3} & 82.5 & 35.7 & 48.9 & \textbf{1.9} & 30.5 & 31.7 & 48.2\\  
			& ESL + IT & 90.2 & 43.9 & \textbf{84.7} & \textbf{35.9} & 28.5 & 31.2 & 37.9 & 34.0 & 84.5 & 42.2 & 83.9 & \textbf{59.0} & 32.2 & 81.8 & 36.7 & 49.4 & 1.8 & \textbf{30.6} & 34.1 & \textbf{48.6}\\
			%         \midrule
			%         Example & Baseline & 0 & 1 & 2 & 3 & 4 & 5 & 6 & 7 & 8 & 9 & 0 & 1 & 2 & 3 & 4 & 5 & 6 & 7 & 8 & 0\\      
			\bottomrule
		\end{tabular}
	}
	\label{tab:gta2cityscapes}
	%\vspace{-0.1cm}
\end{table*}
\begin{table*}[t]
	\centering
	\caption{Comparison of mean intersection-over-union results (in $\%$) for SYNTHIA $\rightarrow$ Cityscapes experiments. mIoU* is the 13-class setup (excluding the classes `wall', `fence' and `pole') as used in earlier works.}%\PPc{mIoU*}
	\resizebox{\textwidth}{!}{%
		\begin{tabular}{l|c|c c c c c c c c c c c c c c c c|c |c}
			\toprule
			\multicolumn{20}{c}{SYNTHIA $\rightarrow$ Cityscapes}\\
			\midrule
			Method & Self-Training & \rotatebox{90}{road} & \rotatebox{90}{sidewalk} & \rotatebox{90}{building} & \rotatebox{90}{wall} & \rotatebox{90}{fence} & \rotatebox{90}{pole} & \rotatebox{90}{light} & \rotatebox{90}{sign} & \rotatebox{90}{veg} & \rotatebox{90}{sky} & \rotatebox{90}{person} & \rotatebox{90}{rider} & \rotatebox{90}{car} & \rotatebox{90}{bus} & \rotatebox{90}{mbike} & \rotatebox{90}{bike} & mIoU & mIoU*\\
			\midrule
			\multirow{3}{*}{AdaptSegNet~\cite{tsai-cvpr2018}} & - & \textbf{83.6} & \textbf{42.0} & \textbf{77.5} & 11.1 & 0.2 & 23.7 & 6.3 & 8.8 & \textbf{78.5} & \textbf{83.1} & 53.7 & 21.1 & \textbf{67.9} & 30.4 & 20.6 & 31.4 & 40.0 & 46.5\\  
			& SSL & 78.9 & 36.6 & 74.8 & \textbf{12.2} & 0.1 & 24.1 & 7.5 & 9.3 & 76.1 & 79.2 & 55.8 & \textbf{22.9} & 66.3 & 29.0 & 24.1 & 38.1 & 39.7 & 46.0\\  
			& ESL & 79.8 & 37.1 & 75.6 & 11.2 & \textbf{0.3} & \textbf{25.0} & \textbf{10.1} & \textbf{9.4} & 77.0 & 82.1 & \textbf{56.1} & 22.2 & 65.2 & \textbf{31.1} & \textbf{25.7} & \textbf{38.7} & \textbf{40.4} & \textbf{46.9}\\  
			%\midrule
			%\multirow{3}{*}{ADVENT} & Baseline & 86.4 & 40.4 & 79.3 & 9.3 & 0.5 & 23.0 & 4.5 & 8.5 & 79.1 & 84.2 & 56.0 & 20.0 & 74.3 & 27.6 & 13.3 & 29.2 & 39.7 & 46.4\\  
			%& SSL & 81.5 & 39.0 & 75.8 & 12.6 & 0.9 & 22.2 & 11.3 & 12.3 & 78.0 & 84.0 & 59.1 & 24.2 & 70.6 & 27.0 & 23.4 & 39.3 & 41.3 & 48.1\\   
			%& ESL & 81.8 & 38.6 & 77.1 & 14.7 & 0.7 & 22.2 & 13.5 & 13.6 & 78.4 & 84.1 & 58.9 & 23.7 & 69.1 & 24.6 & 25.1 & 40.8 & 41.7 & 48.4\\ 
			\midrule
			\multirow{3}{*}{ADVENT~\cite{vu-cvpr2019}} & - & \textbf{85.7} & \textbf{41.4} & \textbf{79.3} & 4.6 & 0.6 & 26.6 & 12.6 & 11.0 & \textbf{79.6} & 84.2 & 53.6 & 21.3 & 70.2 & \textbf{38.9} & 19.2 & 30.7 & 41.2 & 48.3\\  
			& SSL & 84.8 & 40.0 & 78.2 & 8.8 & \textbf{0.8} & 27.4 & 13.2 & 12.4 & 77.6 & \textbf{84.5} & 58.6 & 26.2 & \textbf{75.0} & 36.9 & 20.7 & 44.0 & 43.1 & 50.2\\   
			& ESL & 84.3 & 39.7 & 79.0 & \textbf{9.4} & 0.7 & \textbf{27.7} & \textbf{16.0} & \textbf{14.3} & 78.3 & 83.8 & \textbf{59.1} & \textbf{26.6} & 72.7 & 35.8 & \textbf{23.6} & \textbf{45.8} & \textbf{43.5} & \textbf{50.7}\\  
			%         \midrule
			%         Example & Baseline & 0 & 1 & 2 & 3 & 4 & 5 & 6 & 7 & 8 & 9 & 0 & 1 & 2 & 3 & 4 & 5 & 6 & 7 & 8 & 0\\      
			\bottomrule
		\end{tabular}
	}
	\label{tab:synthia2cityscapes}
	%\vspace{-0.1cm}
\end{table*}

\begin{table}[t]
	\centering
    \caption{Comparison of mean intersection-over-union results (in $\%$) for Mapillary Vistas experiments}
	\resizebox{0.48\textwidth}{!}{%
		\begin{tabular}{l|c|c c c c c c c|c}
			\toprule
			%\multicolumn{2}{c}{} & \multicolumn{8}{|c}{Synthia $\rightarrow$ Mapillary Vistas}\\
			%\midrule
			Method & Self-Training & \rotatebox{90}{flat} & \rotatebox{90}{construction} & \rotatebox{90}{object} & \rotatebox{90}{nature} & \rotatebox{90}{sky} & \rotatebox{90}{human} & \rotatebox{90}{vehicle} &
			mIoU \\
			%\midrule
			%\multirow{3}{*}{AdaptSegNet} & Baseline & 0 & 1 & 2 & 3 & 4 & 5 & 6 & 0 & 0 & 1 & 2 & 3 & 4 & 5 & 6 & 0\\
			%& SSL & 0 & 1 & 2 & 3 & 4 & 5 & 6 & 0 & 0 & 1 & 2 & 3 & 4 & 5 & 6 & 0\\
			%& ESL & 0 & 1 & 2 & 3 & 4 & 5 & 6 & 0 & 0 & 1 & 2 & 3 & 4 & 5 & 6 & 0\\
			\midrule
			\multirow{3}{*}{ADVENT~\cite{vu-cvpr2019}} & - & 87.5 & \textbf{63.2} & 29.2 & 72.2 & \textbf{88.8} & 43.1 & 70.4 & 64.9 \\
			& SSL & 88.3 & 55.4 & 29.1 & 73.3 & 81.8 & \textbf{52.4} & 75.7 & 65.1 \\
			& ESL & \textbf{88.4} & 55.7 & \textbf{32.0} & \textbf{75.4} & 84.3 & 43.5 & \textbf{76.2} & \textbf{65.4} \\  
			%         \midrule
			%         Example & Baseline & 0 & 1 & 2 & 3 & 4 & 5 & 6 & 7 & 8 & 9 & 0 & 1 & 2 & 3 & 4 & 5 & 6 & 7 & 8 & 0\\  
			\bottomrule
		\end{tabular}
	}
	\label{tab:mapillary}
	%\vspace{-0.3cm}
\end{table}

% \begin{table}[t]
%     \centering
%     \caption{Influence of threshold in ESL (ADVENT~\cite{vu-cvpr2019})}   
%     \begin{tabular}{c c}
%         \toprule
%         \multicolumn{2}{c}{GTA5 $\rightarrow$ Cityscapes}\\
%         \midrule
%         Threshold & mIoU \\
%         \midrule
%         Baseline & 43.7\\
%         \midrule
%         0.05 & 45.6\\
%         0.1 & \textbf{45.9}\\
%         0.15 & 45.5\\
%         0.2 & 45.3\\
%         0.3 & 44.9\\
%         Median & 45.1\\
%         \bottomrule
%     \end{tabular}
%   \label{tab:threshold}
% \end{table}

\begin{figure*}[h]
    \vspace{4pt}
	\caption{\small \textbf{Entropy-based pseudo-labels.} The four columns visualize (a) ground-truth label maps, (b) standard pseudo-labels with maximum softmax predictions, (c) our entropy-based pseudo-labels and (d) pseudo-labels in (b) but excluded by our entropy criterion. Most excluded pixels lie in region boundaries where segmentation models are the most uncertain. Experiments show that excluding them boosts the performance of self-trained networks. There are much fewer pseudo-labels added in (c) compared to (b), thus we don't display them.}
	\label{fig:vis}
	\setlength\tabcolsep{1.5pt}
	\footnotesize
	\begin{tabular}{cccc}
		\includegraphics[width=.2352\textwidth]{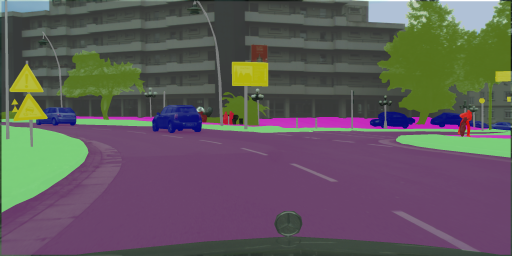} & 
		\includegraphics[width=.2352\textwidth]{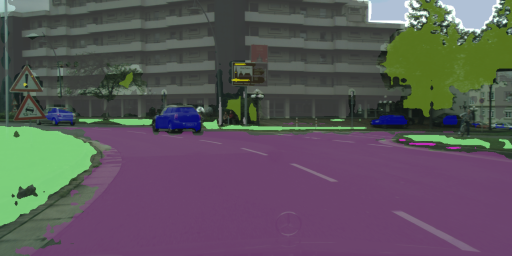} &
		\includegraphics[width=.2352\textwidth]{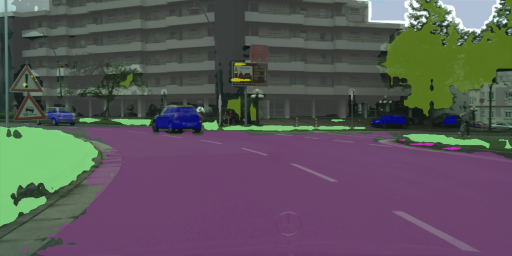} &
		\includegraphics[width=.2352\textwidth]{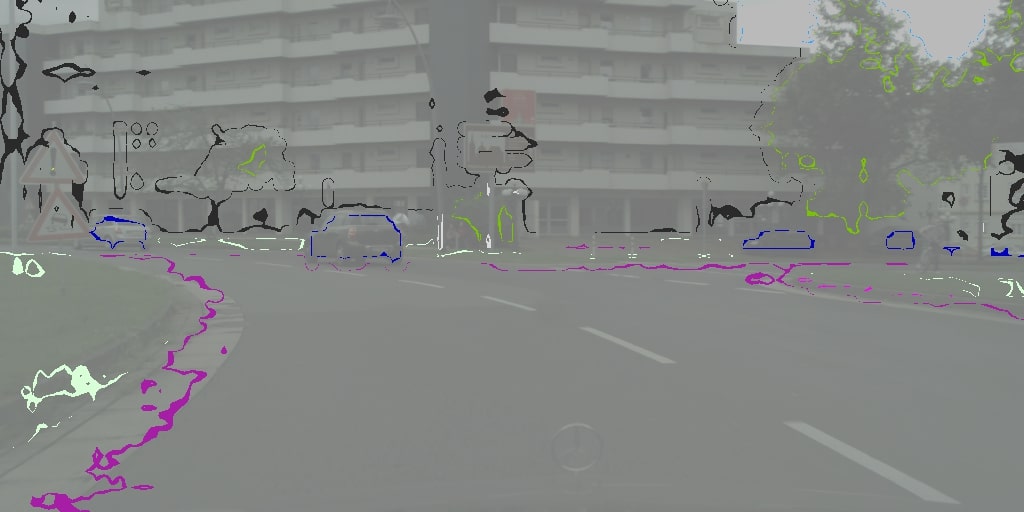} \\
		
		\includegraphics[width=.2352\textwidth]{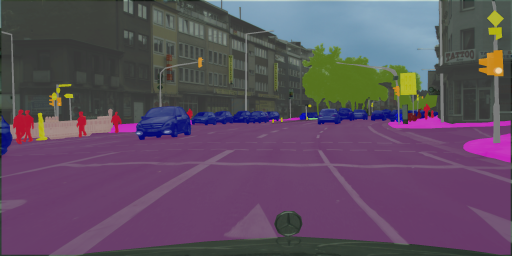} & 
		\includegraphics[width=.2352\textwidth]{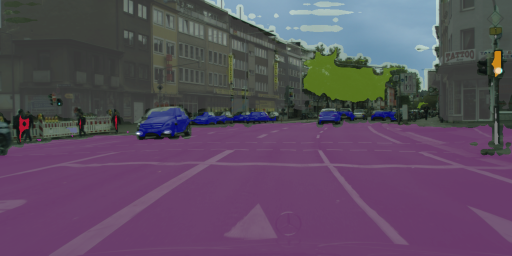} &
		\includegraphics[width=.2352\textwidth]{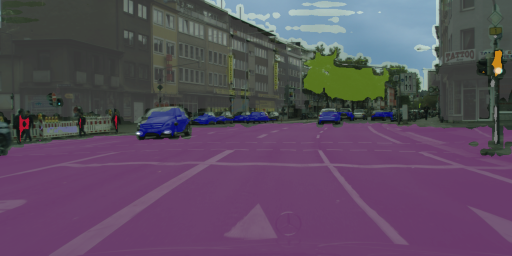} &
		\includegraphics[width=.2352\textwidth]{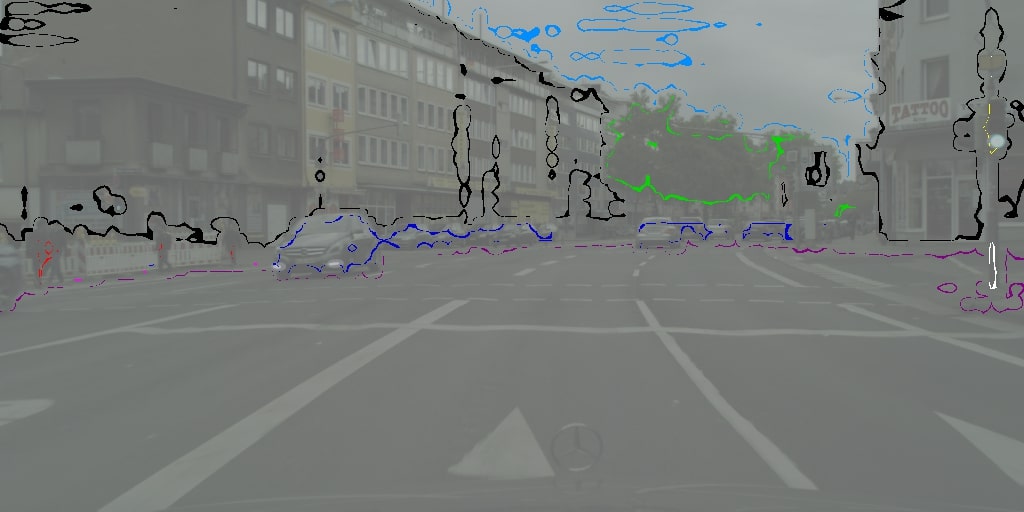} \\
		
		\includegraphics[width=.2352\textwidth]{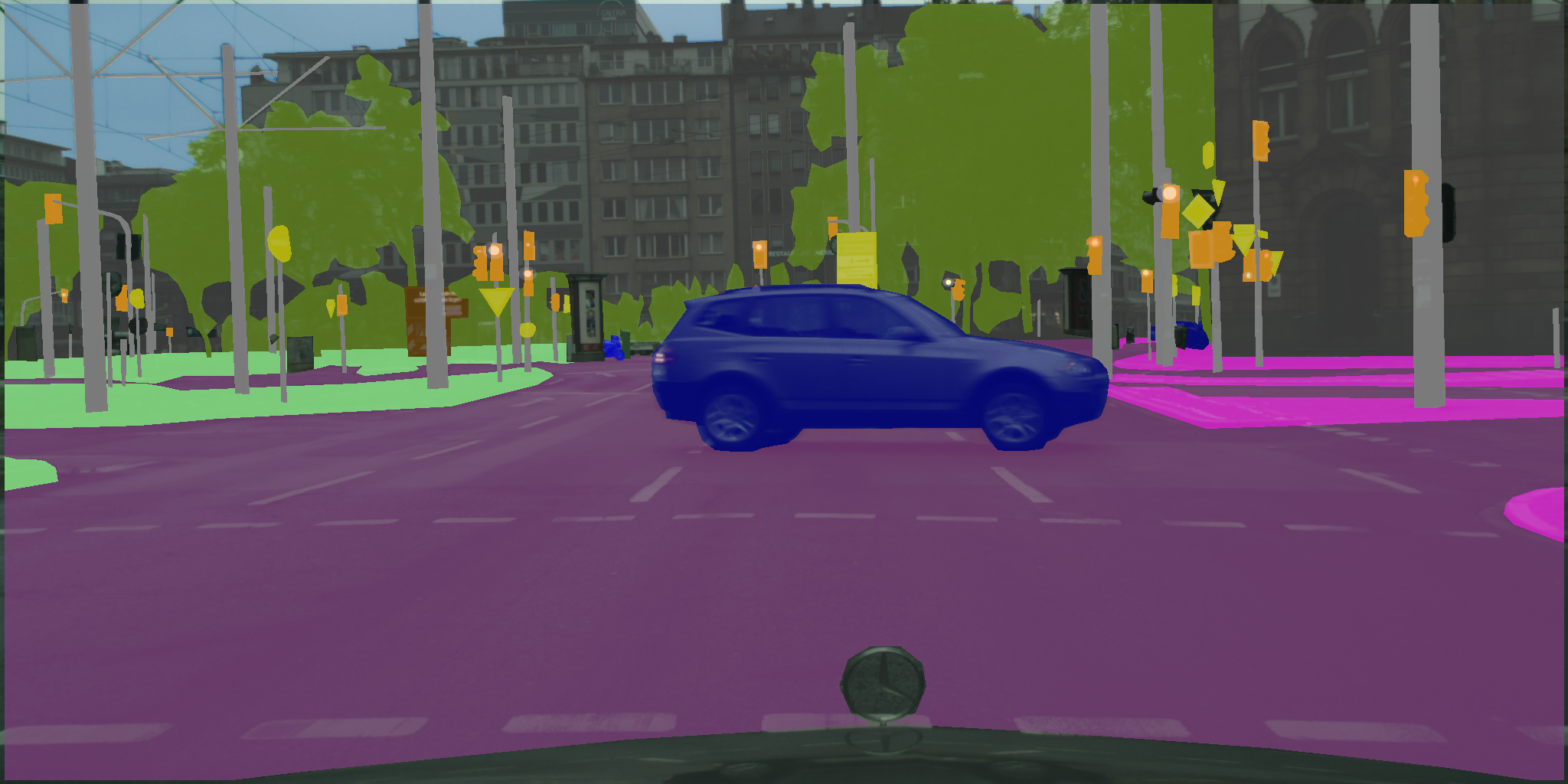} & 
		\includegraphics[width=.2352\textwidth]{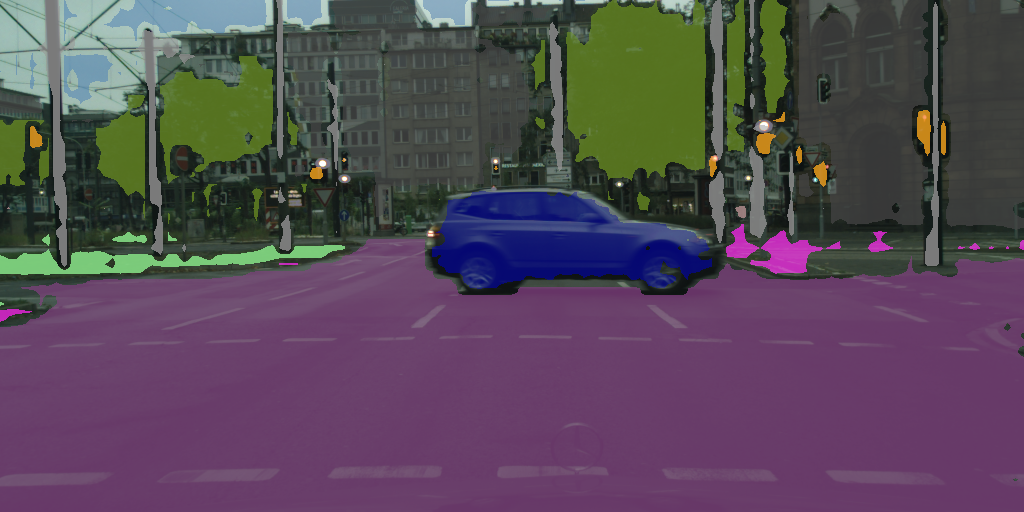} &
		\includegraphics[width=.2352\textwidth]{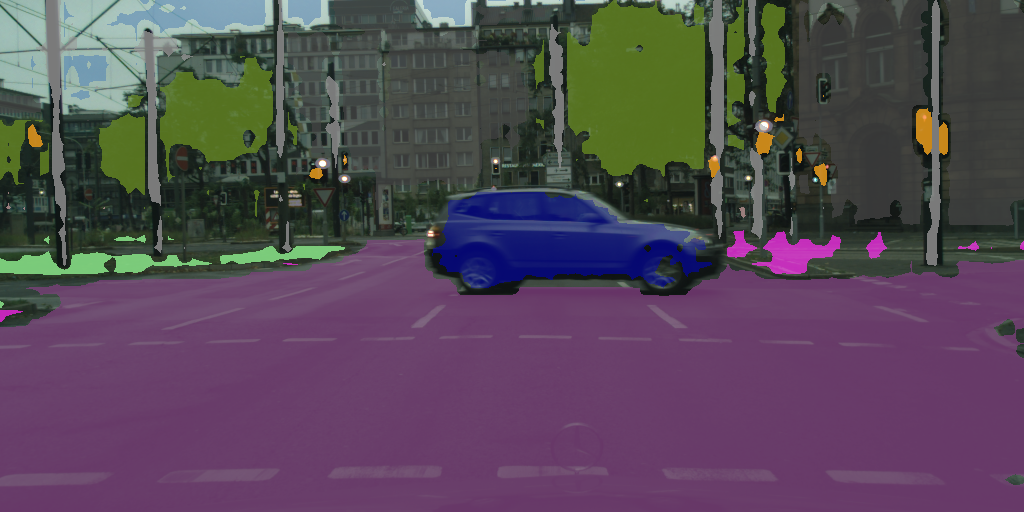} &
		\includegraphics[width=.2352\textwidth]{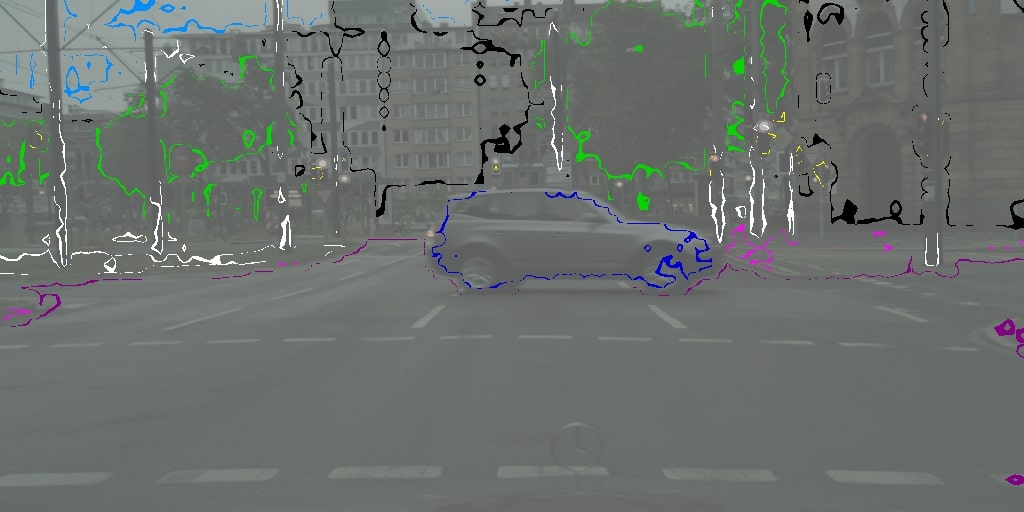} \\
		
		(a) Ground-truth labels & (b) Softmax-based pseudo-labels & (c) Entropy-based pseudo-labels & (d) Excluded by entropy criterion \\
	\end{tabular}
\end{figure*}

\paragraph{GTA5 $\rightarrow$ Cityscapes}
We report in Table~\ref{tab:gta2cityscapes} semantic segmentation performance in terms of mIoU (\%) on the Cityscapes validation set using GTA5 as source domain. As explained in Section~\ref{expdet}, `BDL (step 1)' and `BDL (step 2)' represent the two pretrained models which can be found on the authors' GitHub. We can notice that ESL consistently outperforms SSL on every setup, giving the best performance for each baseline framework. The performance absolute change in terms of mIoU using ESL compared to the baseline state-of-the-art models ranges from $+1.0\%$ to $+2.2\%$ which is a significant improvement.
Along with the quantitative results, Figure~\ref{fig:vis} displays some samples of pseudo-labels extracted with both SSL and ESL. We first note that these maps are very similar within semantic regions.
%, the pseudo-label maps look globally similar inside of the semantic regions between SSL and ESL. 
Indeed, in these regions the softmax prediction is often very peaky with a high maximum score and a low entropy.
%inside of the regions and the entropy is likewise pretty low. 
Nevertheless, marked differences can be observed %some difference may be observed at 
along the boundaries of these regions. These transition areas are those where the prediction models are the most uncertain.
%, despite maximum scores remaining often high.    
%Indeed, the boundaries should be the areas where the model is the most uncertain about its predictions. 
This clearly shows in the fact that most missing pixels in both pseudo-label maps lie on such locations. %those boundary areas.
Around the boundaries however, the prediction entropy tends to get higher even if the softmax prediction score stays high. This results in pixels being rightly excluded from the ESL pseudo-label map while present in the SSL one (column (d) of Figure~\ref{fig:vis}). Reversely, there are much fewer pixels that are added to ESL compared to SSL.

\paragraph{SYNTHIA $\rightarrow$ Cityscapes}

We report in Table~\ref{tab:synthia2cityscapes} semantic segmentation performance in terms of mIoU (\%) on the Cityscapes validation set using SYNTHIA as source domain. Again, ESL consistently outperforms SSL on every setup, even when SSL fails to improve the performance over the baseline.

\paragraph{Mapillary Vistas}
We report in Table~\ref{tab:mapillary} semantic segmentation performance in terms of mIoU (\%) on the Mapillary Vistas validation set using SYNTHIA as source domains and ADVENT as baseline model. Proposed ESL outperforms SSL and improves over the baseline performance.
\begin{table*}[t]
	\centering
	\caption{Comparison of incorrect predictions (in $\%$) selected in the pseudo-labels extracted (ADVENT~\cite{vu-cvpr2019})}    
	\resizebox{\textwidth}{!}{%
		\begin{tabular}{c|c c c c c c c c c c c c c c c c c c c|c}
			\toprule
			\multicolumn{21}{c}{GTA5 $\rightarrow$ Cityscapes}\\
			\midrule
			Method & \rotatebox{90}{road} & \rotatebox{90}{sidewalk} & \rotatebox{90}{building} & \rotatebox{90}{wall} & \rotatebox{90}{fence} & \rotatebox{90}{pole} & \rotatebox{90}{light} & \rotatebox{90}{sign} & \rotatebox{90}{veg} & \rotatebox{90}{terrain} & \rotatebox{90}{sky} & \rotatebox{90}{person} & \rotatebox{90}{rider} & \rotatebox{90}{car} & \rotatebox{90}{truck} & \rotatebox{90}{bus} & \rotatebox{90}{train} & \rotatebox{90}{mbike} & \rotatebox{90}{bike} & \rotatebox{90}{global}\\
			\midrule
			SSL & 19.3 & \textbf{33.5} & 9.8 & 54.0 & 51.3 & \textbf{33.8} & 21.3 & 10.4 & 4.0 & \textbf{37.7} & 6.9 & 4.4 & 33.7 & 2.9 & 53.2 & 50.0 & 66.9 & \textbf{27.5} & \textbf{15.8} & 14.5\\  
			ESL & \textbf{18.5} & 36.1 & \textbf{8.7} & \textbf{53.4} & \textbf{51.2} & 35.1 & \textbf{20.5} & \textbf{10.2} & \textbf{3.4} & 38.7 & \textbf{6.5} & \textbf{4.2} & \textbf{33.1} & \textbf{2.4} & \textbf{53.1} & \textbf{49.8} & \textbf{59.8} & 28.8 & 17.0 & \textbf{13.9}\\
			\midrule
			Rel. Change (\%) & \textbf{-3.9} & +8.0 & \textbf{-11.7} & \textbf{-1.1} & \textbf{-0.3} & +3.9 & \textbf{-3.6} & \textbf{-2.2} & \textbf{-16.1} & +2.6 & \textbf{-5.3} & \textbf{-4.2} & \textbf{-1.8} & \textbf{-18.6} & \textbf{-0.2} & \textbf{-0.5} & \textbf{-10.7} & +5.0 & +7.2 & \textbf{-3.7}\\
			%Diff. & -0.8 & +2.6 & -1.1 & -0.6 & -0.1 & +1.3 & -0.8 & -0.2 & -0.6 & +1.0 & -0.4 & -0.2 & -0.6 & -0.5 & -0.1 & -0.2 & -7.1 & +1.3 & +1.2 & -0.6\\ 
			\bottomrule
		\end{tabular}
	}
	\label{tab:errors}
	%\vspace{-0.3cm}
\end{table*}

\begin{table}[t]
	\centering
	\caption{Influence of threshold $\nu*$ in ESL (ADVENT)}   
	\resizebox{.48\textwidth}{!}{%
		\begin{tabular}{c | c | c c c c c c}
			\toprule
			\multicolumn{8}{c}{GTA5 $\rightarrow$ Cityscapes}\\
			\midrule
			Threshold & Baseline & 0.05 & 0.1 & 0.15 & 0.2 & 0.3 & Median \\
			\midrule
			mIoU & 43.7 & 45.6 & \textbf{45.9} & 45.5 & 45.3 & 44.9 & 45.1\\
			\bottomrule
		\end{tabular}
	}
	\label{tab:threshold}
	%\vspace{-0.5cm}
\end{table}
\subsection{Ablation Studies}
\label{sec:ablation}
\paragraph{Choice of threshold $\nu^*$} Let us describe how to choose the threshold $\nu^*$ such that 
% filter out data in order to get 
a good balance is achieved between having as many high confidence predicted labels as possible and avoiding as much as possible noise from incorrect predictions. A quick computation suggests $0.1$ as a good threshold value. Indeed, considering the maximum softmax prediction score for a given pixel is $0.95$ on a 19-classes setup, the entropy of the distribution would range from $0.07$ in the best case to $0.12$ in the worst case. We confirm this choice experimentally. We show in Table~\ref{tab:threshold} segmentation results on the domain adaptation problem GTA5~$\rightarrow$~Cityscapes with the ADVENT baseline using different thresholds in the ESL label extraction. Alternately, we show the limit case where the threshold is always selected as the median of the entropy for each class. The result of this experiment is shown on the last row of  Table~\ref{tab:threshold}. When the threshold is higher than $0.1$, the incorrect predictions degrade the quality of the pseudo-label maps and induce more noise in the training. When the threshold is lower than $0.1$, we don't keep as many confident predictions, slightly reducing the effectiveness of the pseudo-labeling. This experiment confirms that $0.1$ seems to be a good threshold for ESL.

\paragraph{Incorrect predictions in the pseudo-labels} In order to further motivate the choice of entropy as a confidence measure in the proposed ESL method over the softmax prediction score of SSL, we report in Table~\ref{tab:errors} the ratio of incorrect predictions (in $\%$) in the selected pseudo-labels for every class for both SSL and ESL on the GTA5 $\rightarrow$ Cityscapes experiment with the ADVENT baseline (the lower the score, the better). Additionally, the last row of the Table displays the relative change in the ratio of incorrect predictions for every class. The results show that ESL performs significantly better than SSL on the easier-to-predict classes (less than $10\%$ incorrect predictions in the selected pseudo-labels). Indeed, ESL reduces the number of incorrect predictions for those classes by a significant margin, ranging from $4.2\%$ up to $18.6\%$. This means that ESL induces significantly less noise in the training for those classes. These changes can be explained by the ``overconfidence'' the model may have in terms of softmax prediction score for easier-to-predict classes, which is not as significant for the entropy as measure of confidence. Overall, ESL decreases the ratio of incorrect predictions for 14 classes out of 19 and globally reduces the ratio of incorrect predictions by $3.7\%$ over SSL.

\section{Conclusion}
In this paper, we propose an Entropy-guided Self-supervised Learning strategy for semantic segmentation adaptation problems. We demonstrate with a variety of experiments that this method consistently improves the performance of state-of-the-art domain adaptation models on synthetic-to-real problems and outperforms standard softmax-based self-supervised learning approaches. 

{\small
\bibliographystyle{ieee_fullname}
\bibliography{bibliography}
}

\end{document}